\documentclass[preprint,12pt]{elsarticle}

\usepackage{amssymb}
\usepackage{svg}
\usepackage{url}
\usepackage{subcaption}
\usepackage{multirow}

\journal{Pattern Recognition}

\begin{document}

\begin{frontmatter}



\title{Why Divisive Normalization works in image segmentation?}


\author[label1]{Pablo Hernández-Cámara\corref{cor1}}
\author[label1]{Jorge Vila-Tomás}
\author[label1]{Paula Dauden-Oliver}
\author[label2,label1]{Nuria Alabau-Bosque}
\author[label1]{Valero Laparra}
\author[label1]{Jesús Malo}

\affiliation[label1]{organization={Image Processing Lab, Universitat de València},
                     city={Paterna},
                     postcode={46980},
                     country={Spain}}
\affiliation[label2]{organization={ValgrAI: Valencian Grad. School Research Network of AI},
                     city={València},
                     postcode={46022}, 
                     country={Spain}}
\cortext[cor1]{pablo.hernandez-camara@uv.es}
\cortext[cor2]{Work accepted at Neurcomputing, DOI: 10.1016/j.neucom.2025.130569}

\begin{abstract}

Autonomous driving is a challenging scenario for image segmentation due to the presence of uncontrolled environmental conditions and the eventually catastrophic consequences of failures. Previous work suggested that a biologically motivated computation, the so-called Divisive Normalization, could be useful to deal with image variability, but its effects have not been systematically studied over different data sources and environmental factors.

Here we put segmentation U-nets augmented with Divisive Normalization to work far from training conditions to find where this adaptation is more critical.  We categorize the scenes according to their radiance level and dynamic range (day/night), and according to their achromatic/chromatic contrasts. We also consider video game (synthetic) images to broaden the range of environments. We check the performance in the extreme percentiles of such categorization. Then, we push the limits further by artificially modifying the images in perceptually/environmentally relevant dimensions: luminance, contrasts and spectral radiance. Results show that neural networks with Divisive Normalization get better results in all the scenarios and their performance remains more stable with regard to the considered environmental factors and nature of the source.

Finally, we explain the improvements in segmentation performance in two ways: (1) by quantifying the invariance of the responses that incorporate Divisive Normalization, and (2) by illustrating the adaptive nonlinearity of the different layers that depends on the local activity.

\end{abstract}


\begin{highlights}
\item We test the generalization ability of segmentation models to environmental data changes.
\item Bio-inspired model gets better segmentation results in all scenarios.
\item Divisive Normalization produces higher gains in the low luminance and contrast domains.
\item Bio-inspired model performance is more stable across different
factors and data sources.
\item Bio-inspired models are more invariant to data diversity.
\end{highlights}

\begin{keyword}
Divisive Normalization \sep Segmentation \sep U-Net \sep Invariance \sep Generalization \sep Adaptation \sep Autonomous Driving

\end{keyword}

\end{frontmatter}


\section{Introduction}
\label{sec_introduction}

Autonomous driving includes scenes at \emph{any} time of the day and night, may include wild weather conditions, shadows, and illumination vary at different time scales. These variations lead to high dynamic range scenes both in luminance and color: imagine bright streetlights in the night~\cite{fairchilddatabasehdr} and saturated colors in sunny scenes as colorfulness increases with luminance~\cite{fairchild2013color}. Environmental changes modify average luminance and contrast (e.g. in fog or snow)~\cite{fog_cityscapes} and induce non-additive noise (e.g. rain)~\cite{raincityscapes}. Moreover, radiance level and spectral illumination change along the day~\cite{Camps11,Jimenez13}, and induce nontrivial contrast changes in otherwise continuous objects (e.g. shadows~\cite{fairchild2013color, Cavanagh18}, or interreflections~\cite{laparra2012nonlinearities,deeb2018interreflections}).
This uncontrolled scenario implies that similar reflectances and textures lead to highly variant measurements in the camera. The possible fatal consequences of mis-segmentation in autonomous driving~\cite{Tesla} imply that models must have an excellent performance despite all the above variabilities, i.e. they have to be invariant under these changes.
Moreover, the complexity in these scenes is not easy to simulate in synthetic scenarios, so if computer-generated images were used to train (data-hungry) algorithms~\cite{saleh2018effective,ros2016synthia}, they could generalize poorly in real life.

\subsection{Simple illustration of the diversity problem}

\begin{figure}[]
\centering
  \includegraphics[width=1.0\textwidth]{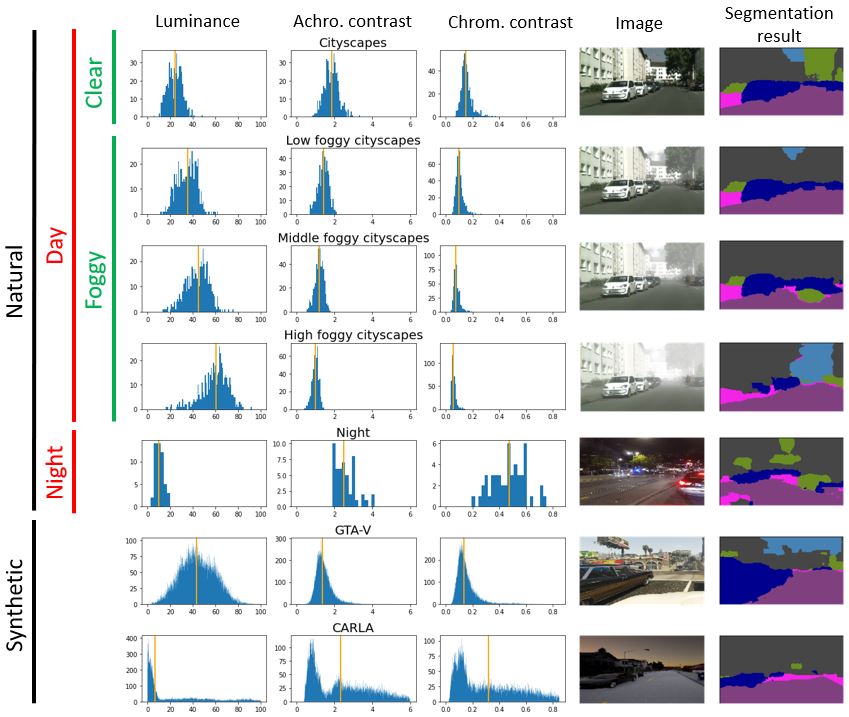}\\
  \caption{ {\footnotesize \textbf{Motivation: colors and the energy of visual textures change with the environment and data nature and highly affect the segmentation results.} Histograms of luminances (left histogram column), achromatic contrast (middle histogram column) and chromatic contrasts (right histogram columns) for 7~different datasets (rows). Vertical orange lines in the histograms represent their median values.
  The definition of contrasts (energy of spatial modulation of luminance and color) is described in Section~\ref{results_global_extreme}. Last two columns show an example of each dataset and the segmentation result of a U-Net model~\cite{ronneberger2015u} trained with natural daytime and clean images.
  The first row shows the distributions corresponding to the natural, daytime scenes from Cityscapes~\cite{Cordts2016Cityscapes}. 2nd to 4th rows correspond to the same scenes modified to include different fog levels~\cite{fog_cityscapes}. 5th row corresponds to real urban night images~\cite{daytime_2_nighttime}. 6th row corresponds to scenes from the famous video game GTA-V~\cite{gta} and last row corresponds to computer-generated scenes using the virtual-reality framework CARLA~\cite{Carla_dataset}.}}\label{datasets_statistics}
\end{figure}

While the above diversity in the images depends on complex interactions in the scenes and on its natural or sythetic origin, a simple univariate analysis of relevant visual descriptors 
can illustrate the problem. 
Figure~\ref{datasets_statistics} includes 7~natural and synthetic datasets of driving scenes in different environments.
We analyze these scenes according to their \emph{luminance}, and to the \emph{achromatic} and \emph{chromatic contrasts} of their textures.
Histograms show how diverse these descriptors are.
Figure~\ref{datasets_statistics} also includes a segmentation example from each dataset and the negative effect of this diversity when using a segmentation algorithm trained only with real daytime images. Note how the segmentation result becomes worse as the fog level increases, the mis-detection of the car in the night scenario and the mis-detection of a (clearly synthetic) sky in the CARLA scenario.

Description of the scene conditions according to these visual features (defined below in Section~\ref{results_global_extreme}), is of course an oversimplification. However, note the clear patterns of change when changing the conditions.
For instance, fog increases mean luminance: see how the peak in the first column (rows 1-to-4) moves to the right with the fog level. Fog reduces the achromatic contrast (see the central histogram column, rows 1-to-4) because darker objects become lighter due to the reflected light via scattering. This reduces the energy of spatial modulations of luminance thus impeding texture-based segmentation. Fog also reduces color saturation (and hence chromatic contrast in the right column) because back-scattered light is white (or broadband). 
This reduces the energy of spatial modulations of color thus impeding segmentation based on chromatic textures. Night scenes (5th row) are obviously darker (left), but interestingly, contrasts (both achromatic and chromatic) are substantially higher (bright streetlights, saturated neon adds...). Finally, completely synthetic (video-game-like) data shows distributions that deviate from natural scenes, as illustrated by the last two rows of Figure~\ref{datasets_statistics}.
On the one hand, note that the luminance peak in CARLA~\cite{Carla_dataset} is darker than real night images in~\cite{daytime_2_nighttime}, while both constrasts are wider than in natural databases~\cite{Cordts2016Cityscapes,daytime_2_nighttime}. On the other hand, GTA-V~\cite{gta} displays a larger luminance range than the natural scenes in~\cite{Cordts2016Cityscapes} (day), and~\cite{daytime_2_nighttime} (night). Note that the luminance peak in GTA-V is wider than the natural distributions combined.

\subsection{Approaches to cope with diversity}

Proper consideration of the mentioned variability for successful segmentation is a matter of active research. For instance, authors from~\cite{daytime_2_nighttime} train 
series of segmentation models for different periods of time from daylight to nighttime through the sunset in order to cope with illumination variability.
In particular, they use the model trained at certain time as starting point for the model valid for the next period of time, so that the resulting model could be invariant under illumination changes. 
In fact, the rationale of adding synthetic weather conditions to natural scenes~\cite{fog_cityscapes, raincityscapes} is taking into account such variability through data augmentation.
However, to train models over the widest range of conditions one may need completely synthetic data, as in~\cite{gta}.
However, as illustrated in Figure~\ref{datasets_statistics}, synthetic data may have different statistics and its use may lead to the conventional out-of-distribution problem.

As opposed to (computationally demanding) data augmentation, a recent bio-inspired layer for image segmentation has been proposed to (intrinsically) cope with this kind of data variability~\cite{Ortiz_2020_CVPR, HERNANDEZCAMARA202364}, the so-called Divisive Normalization of visual neuroscience~\cite{Carandini2012NormalizationAA, NormalizationPrinciplesinComputationalNeuroscience}. Both works~\cite{Ortiz_2020_CVPR, HERNANDEZCAMARA202364} use the challenging scenario of autonomous driving as a useful case study. Divisive Normalization may be good for invariant segmentation because the response of each neuron is normalized by the responses of other neurons tuned to neighbour locations and features, so each response is adapted to the local activity.
The authors of~\cite{Ortiz_2020_CVPR} focused on the difference with regular batch and layer normalization. In particular, they analyzed the effect of the size and shape of the neighbourhoods in the normalization. However, while they report gains in a specific segmentation task, they do not check the invariance to environmental nor data diversity.  On the contrary, the proposal in~\cite{HERNANDEZCAMARA202364} was focused on the eventual invariance to atmospheric factors. In particular, results showed potential benefits when facing achromatic contrast reduction due to different fog levels, suggesting that the proposed layer could induce more general invariances. In~\cite{Ortiz_2020_CVPR, HERNANDEZCAMARA202364} the environment was not changed in any controlled way and in~\cite{HERNANDEZCAMARA202364} fog was the only illumination factor. Moreover, the nature/quality of the data (real vs synthetic) was not considered in these seminal works. 

\subsection{Scientific questions addressed}

Divisive Normalization may be useful for image segmentation, but given the obvious variability of measurements (colors and visual textures) in the wild (Fig.~\ref{datasets_statistics}), natural questions arise:

\emph{How much does the Divisive Normalization improve the segmentation performance? Does it help in other conditions such as different weather, illumination, and data quality? 
And, if so, why does it improve the segmentation in these specific conditions?}

Detailed responses to these questions (e.g. in systematically partitioned regions of the problem) are important because aggregate performance metrics and lack of access to fine-grain results limit understanding of the solutions~\cite{Orallo23}.
Evaluation of \emph{capabilities} rather than \emph{task performance} or predicting the places where the system is going to gain/fail is important when safety is an issue~\cite{zhou2023}.
In this regard, explainability given by certain analytical computations such as the proposed Divisive Normalization is a good complement to pure maximization of the performance measure~\cite{Martinez19}. And quantitative descriptions of the reasons for improvement in the task (e.g. invariance of the representation) are highly desirable too. 

This work answers these questions which were not addressed in~\cite{Ortiz_2020_CVPR, HERNANDEZCAMARA202364}.
Here we systematically analyze and control different environmental factors (luminance, energy of the achromatic and chromatic textures, and spectral illumination) and the quality of the images (natural and synthetic) to test the effect that Divisive Normalization has in segmentation networks when they face that diversity. 
This allows us to delineate the conditions where the use of the proposed layer is more critical.
Finally, we explain the benefits of the considered layer through quantitative measures of the invariance of the prediction space, and an illustrative analysis of the adaptive nonlinearities.

\vspace{0.2cm}

The rest of the paper is organized as follows: 
First, in section \ref{sec_div_norm} we review the formulation of Divisive Normalization. Second, in section \ref{sec_data_models} we develop the different models (neural networks) and data (scenes to segment) that we use in our study as well as the different environmental factors considered and, following Figure~\ref{datasets_statistics}, how we systematically change them. Next, in section \ref{sec_experiments_results} we expose the results that we obtain over different data diversity factors and over the different controlled changes we introduce. Then, in section \ref{analysis_section} we analyze where the advantage of the Divisive Normalization comes from and how it is achieved and finally we conclude in section \ref{sec_conclusion} with an example of use showing the better segmentation of the models with Divisive Normalization.

\section{Background on Divisive Normalization}
\label{sec_div_norm}

Divisive Normalization \cite{heeger1992normalization} is the canonical computation that accounts for adaptation in biological neurons~\cite{Carandini2012NormalizationAA,NormalizationPrinciplesinComputationalNeuroscience}. It is a local normalization in which the response of a sensor is normalized not only by its own value but also by taking into account the values of its local surroundings: 

\begin{equation}
    y_k = \frac{z_k}{( \beta_k + \sum{\gamma_{k,s} * |z_s|^{\alpha_s}}_s)^\epsilon_k}
\end{equation}

where the linear response of a sensor, $z_k$, is inhibited (normalized) by a pool of the activity of neighbour sensors. In the context of computer vision models that work with images as input, it implies that one pixel is modulated not only by its own value but also by the values of close pixels. Note that the division is a point-wise operation, and the sum in the denominator represents the interaction between the considered pixel and its neighbourhood. The exponents $\epsilon$ and $\alpha$ control the norm of the pool. The ratio between the constant $\beta$ and the convolution kernel $\gamma$ determines the level of non-linearity. The interaction kernel in the denominator, $\gamma$, can have any arbitrary structure to take into account the surroundings but here we build it to have the form of a usual spatial convolution with dense connections between the input channels.

\section{Models, data and environmental factors}
\label{sec_data_models}

\subsection{Models}
\label{sec_models}

\begin{figure*}[h]
\centering
  \includegraphics[width=1.0\textwidth]{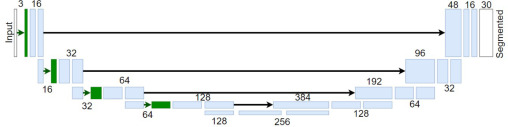}\\
  \caption{\textbf{U-Net for segmentation using four Div. Norm. layers (4-DN)}. Divisive Normalization layers are indicated in green. Numbers above the layers indicate the number of features in each block and black arrows represent the skip unions. The no-DN model does not have any of the green layers, i.e. it does not have Divisive Normalization layers. The consideration of the DN layers only increases the number of parameters by 1.8\% with regard to the no-DN model. Image from \cite{HERNANDEZCAMARA202364} reproduced with author permission.}
  \label{models}
\end{figure*}

Building the Divisive Normalization non-linearity in an automatic differentiation environment allows us to use it and optimize its parameters in any machine learning model\footnote{https://github.com/pablohc97/TFM/blob/main/GDN.py}. Following \cite{HERNANDEZCAMARA202364} ablation study, we build two different models, one with four Divisive Normalization layers (called 4-DN) and the other without any (no-DN), both U-Net like \cite{ronneberger2015u}. More particularly, Figure \ref{models} shows the exact structure of our models and where the Divisive Normalization is included. 

We built our models as in \cite{HERNANDEZCAMARA202364} to be able to compare with their results. Therefore, in our models, $\gamma$ corresponds with a $3 \times 3$ convolution kernel so that the close neighbours around the considered pixel are taken into account. The $\beta$ parameter is also trained, but we set $\epsilon$ and $\alpha$ to $1$ which makes the training more stable.

To quantify the models' performance, we use the commonly used segmentation metric Intersection over Union (IoU). It measures the overlap between the predicted segmentation mask and the real ground truth.

\subsection{Data}
\label{sec_data}

In our experiments, we used four different datasets to train and test the models and assess the Divisive Normalization effect. 

We used Cityscapes Dataset \cite{Cordts2016Cityscapes}, one of the most famous semantic segmentation datasets for autonomous driving. It includes scenes with 30 classes annotated segmentation ground truths. Images were taken in good weather conditions and during the day by a car on the streets of 50 cities in Germany. 

In addition, we use the Nighttime Driving-test dataset \cite{daytime_2_nighttime}. It contains 50 segmented real images from Swiss cities taken at night. This dataset will allow us to test the Divisive Normalization good effect when facing low-luminance images.

\begin{figure*}[b]
\centering
  \includegraphics[width=1.0\textwidth]{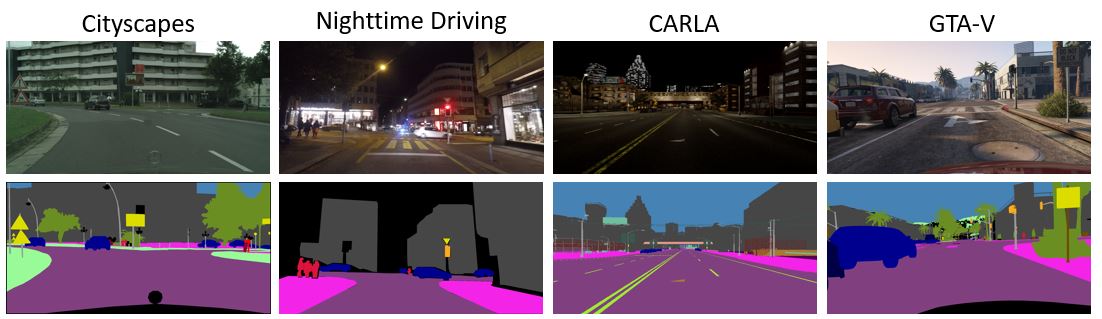}
  \caption{\textbf{Example images and segmentation ground truth of the datasets}. From left to right an image of Cityscapes, Nighttime Driving, CARLA Simulator and GTA-V.}
  \label{datasets}
\end{figure*}

The previous datasets are made of real images. However, generating real datasets is very time-consuming because we need to manually generate the ground truths and they do not cope with all the possible data variability that models can find in real life. For these reasons, synthetic datasets are rising in popularity. Following this idea, we also used two synthetic datasets. First, we used the CARLA Simulator (Car Learning to Act) \cite{Carla_dataset}. It is an open simulator for urban driving, developed as an open-source layer over Unreal Engine 4. We generate a new dataset that contains 20000 images of all weather (sunny, rainy and foggy) and time conditions (morning, day, evening and night) from two different simulated cities with their corresponding segmentation ground truth \cite{IPL_Carla_dataset}.
We divide it into $80\%$, $10\%$ and $10\%$ to train, validate and test, which gives us $16000$, $2000$ and $2000$ train, validation and test images.

Finally, we used the GTA-5 dataset \cite{gta}. It contains almost 25000 synthetic images rendered using the open-world video game Grand Theft Auto 5. Images are from the car perspective in the streets of American-style virtual cities and they also include different weather (sunny and rainy) and time (day, evening and night) conditions.

To summarize all the datasets, figure~\ref{datasets} shows an image of the four datasets and their associated segmentation ground truth, which we want to predict with our models.

\subsection{Environmental factors}

To test how different environmental factors affect the models and to check the effect of Divisive Normalization in controlled experiments, we have to characterize and modify the visual appearance of the images. 
To do so, we follow the intuition introduced in Figure~\ref{datasets_statistics}.

\begin{figure*}[b]
\centering
  \includegraphics[width=1.0\textwidth]{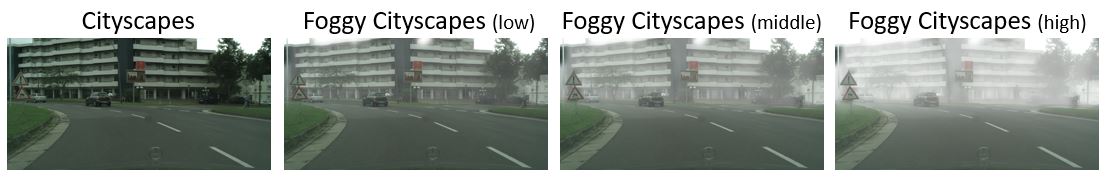}
  \caption{\textbf{Example of fog severities}. An image from Cityscapes and their corresponding versions in Foggy Cityscapes with different severities.}
  \label{foggy_dataset}
\end{figure*}

We first consider Foggy Cityscapes \cite{fog_cityscapes}. It is a synthetic foggy dataset that simulates fog on the real scenes of Cityscapes. Each Cityscapes image is rendered from a clear image and a depth map to include fog of controlled (low, medium and high) severity. Each level is characterized by a constant attenuation coefficient (0.005, 0.1 and 0.2) corresponding with a visibility range of 600, 300 and 150 meters respectively. Therefore, this dataset has the same Cityscapes images but with simulated fog as shown in figure \ref{foggy_dataset}.

In order to push the models to the limit, we test them on the more difficult or extreme scenarios, i.e. in the images of the datasets which have extreme visual appearance according to simple descriptors as those illustrated in Figure~\ref{datasets_statistics}.
To do so, we create different partitions of the real and synthetic data (Cityscapes and GTA-V) according to the extreme values of luminance, achromatic contrast and chromatic contrast. 
The rationale to propose these partitions is challenging the models that may solve the segmentation problems using color or texture variations in certain luminance levels.

The methodology to identify extreme subsets is the following. We first express the real and synthetic images in a classical Achromatic Tritanopic (red-green) and Deuteranopic (yellow-blue), or ATD, color space~\cite{fairchild2013color}. Once in this color space, we can easily define the mean luminance, $\mu_A$, and the achromatic and chromatic contrast, $C_a$ and $C_{chro}$ as:
\begin{eqnarray}
    C_{a} & = & \sqrt{2} \cdot \frac{\sigma_{A}}{\mu_{A}}\\
    C_{chro} & = & \sqrt{2} \cdot \frac{\sqrt{\sigma_{T}^{2} + \sigma_{D}^{2}}}{\mu_{A}}
\end{eqnarray}

\noindent where $\mu_A$ is the spatial mean of the achromatic channel A, and $\sigma_{i}$ correspond to the standard deviation of channel ${i}$ over the spatial extent of the image.
The above descriptors are applicable to natural images but are consistent with the standard definition of Michaelson contrast for sinusoidal gratings of luminance~\cite{Peli90}, and with the definition of perceptual colorfulness~\cite{fairchild2013color}. 
These contrasts represent the amplitude (or energy) of the achromatic and chromatic textures in the image.
Once we have $(\mu_A, C_a, C_{chro})$ for each image, we sort them accordingly and we identify the subsets which are in the 15th/20th and 85th/80th percentile of each visual feature for the GTA-V and Cityscapes datasets. 
This determines six new partitions: low mean luminance (called Low A), high mean luminance (High A), low achromatic contrast (Low A ctr), high achromatic contrast (High A ctr), low chromatic contrast (Low chrom ctr) and high chromatic contrast (High chrom ctr).

\begin{figure}[b]
\centering
  \includegraphics[width=1.0\textwidth]{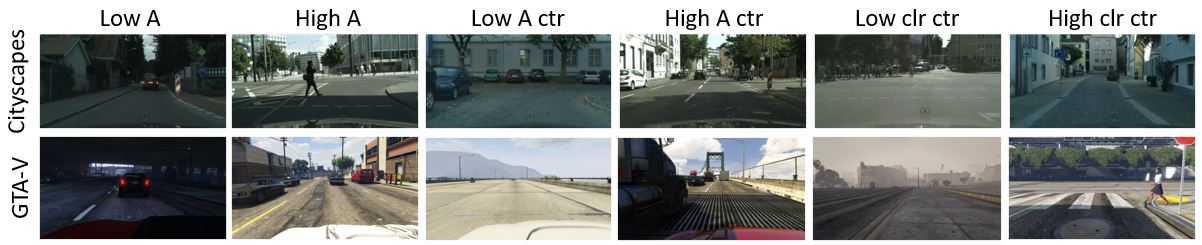}\\
  \caption{\textbf{Representative extreme images of the dataset partitions}. From left to right it shows the images with the lowest and higher mean luminance, achromatic contrast and chromatic contrast.}\label{example_partitions}
\end{figure}

Figure \ref{example_partitions} shows the extreme images of these six partitions for both datasets. Note that low mean luminance corresponds with very dark images while high mean luminance images are really bright. On the achromatic contrast partitions, the lowest ones are images that are almost flat, texture-less scenes, while the high achromatic contrast images have both really dark and bright sections. In the low chromatic contrast images, we find images without almost any color but in the high chromatic contrast images there are more colourful images over different backgrounds.

Although, as seen in the results section, these dataset partitions and the foggy scenes help us to test the model in extreme conditions and they are useful to get initial ideas of where the critical situations are, these subsets are limited in different ways.  
First, the values of the descriptors for these subsets cannot be controlled in a smooth way, and second, the range of the visual descriptors is limited by the set of images already available in the original datasets.

In order to overcome these limitations and extend the range of our exploration in a systematically controlled way, we artificially modify the original images in five relevant visual dimensions: the already mentioned 
(1)~mean luminance, (2)~achromatic contrast, (3)~chromatic contrast, and the spectral illumination, which amounts to (4)~hue angle, and (5)~saturation.

\begin{figure}[t]
\centering
  \includegraphics[width=1.0\textwidth]{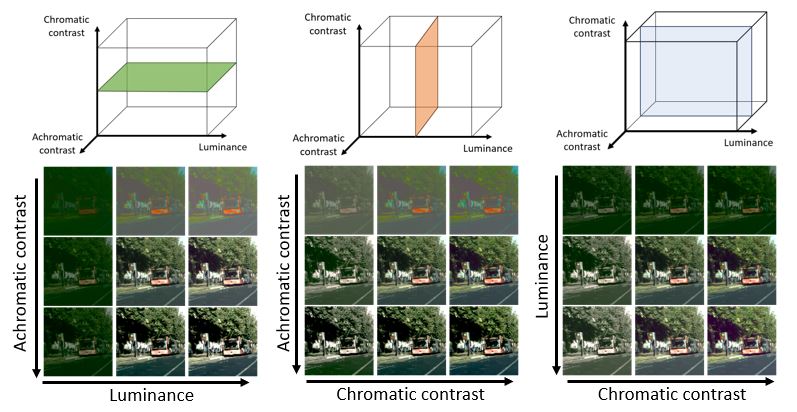}\\
  \caption{\textbf{Controlled modification of luminance and contrasts}. Slices and examples from the 3D tensor (luminance, achromatic contrast
and chromatic contrast). Note that the central image represents the original image. Figure shows 3 values/dimension but the actual intervention took 10 linearly spaced values/dimension.}\label{tensor_example}
\end{figure}

First, we address the variation of the three descriptors introduced in Figure~\ref{datasets_statistics}. 
We select 100 images from the Cityscapes test dataset and manually increase and reduce their mean luminances, achromatic contrasts and chromatic contrasts in turns in the ATD space.  
In this way, we generated a tensor of 3 dimensions $(\mu_A, C_a, C_{chro})$ where in each point we have 100 Cityscapes images with fixed luminance, achromatic contrast and chromatic contrast. We modified the original values of the descriptors by factors from 0.5 to 1.4 and the set is available here~\cite{IPL_City_LuminanceContrast}. 
Figure~\ref{tensor_example} illustrates three different slices from the tensor, fixing the constant dimension to its original value.

Second, we can introduce variations in spectral illumination in order to shift the colors in the scene in a systematic way.
To do that, we use the following \emph{approximated but convenient} approach. We generate a white (equienergetic) spectrum and find the lambertian reflectance of each pixel so that when the white light is reflected, we obtain the closest tristimulus values in each pixel.
Once we have the reflectance of each pixel, we can generate light spectrums of different dominant wavelengths (different hues) and saturations and use them as illuminats for these reflectant scenes. In particular we select illuminants in a regular 2D-grid in polar coordinates around the white point in the CIE xy 1931 color diagram for 20 hue angles and 6 saturation distances (radius) from the white point.

\begin{figure}[h]
\centering
\includegraphics[width=\linewidth]{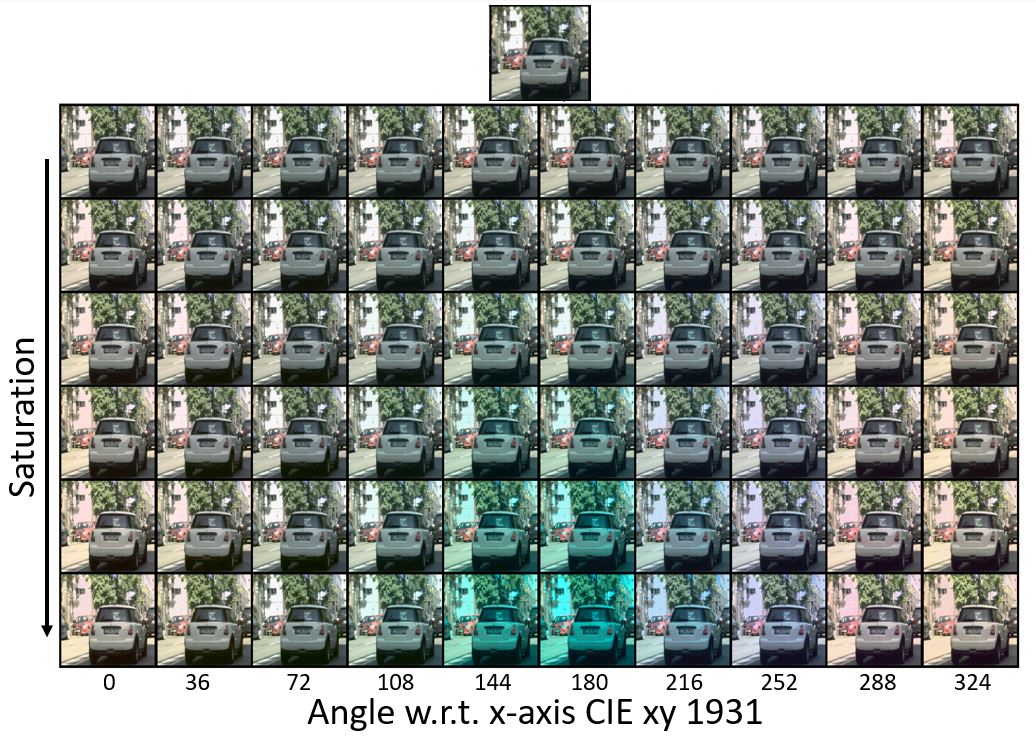}\\
  \caption{\textbf{Controlled change of spectral illumination}. The top image is one original image and the bottom matrix shows the obtained images from different illuminant hues (orientation angle in CIE xy diagram) and saturations (distances to the white point).
  This illustration shows 10 hues but the intervention considered spectra of 20 linearly spaced hue angles.}
\label{illuminant_changed_example}
\end{figure}

This approach takes part of the physics of color into account but obviously makes gross approximations.
First, it disregards complex mutual illuminations in the unknown geometry of the scenes. We worked on models of interreflections and we know that they give rise to nontrivial distributions of tristimulus values~\cite{laparra2012nonlinearities,deeb2018interreflections}. And these do not appear in scenes assumed to be a flat lambertian mosaic. 
Second, the tristimulus-to-spectrum transform is not univocal because of the strong dimensionality reduction in the spectrum-to-tristimulus transforms, or metamerism~\cite{Stiles00}. 
Therefore, the hack based on looking for the best matching reflectance for each pixel in the Munsell database to minimize tristimulus error (as done in Colorlab~\cite{Colorlab02}) is just one of other possible solutions to build the hyperspectral scene from the tristimulus scene.

Actual changes of spectral illumination in real scenes will lead to images which will differ from the example shown in Figure~\ref{illuminant_changed_example}.
However, reasonable visual aspect of the result justifies the assumption of the approximations 
because of the benefits we get in extending the databases in a controlled way in order to check the color constancy of the segmentation.
The final dataset of 100 images with $20\times6 = 120$ modified illuminants is available here~\cite{IPL_City_Illuminants}.

Table \ref{table_datasets} shows a summary of the different datasets and modifications and how many images they have that we use to train, validate and test the different models.

\textbf{\begin{table}[h]
\caption{Summary of the number of training, validation and test images per dataset.}
\label{table_datasets}
\centering
\begin{tabular}{c|c|c|c}
Dataset & Train images & Val. images & Test images \\\hline
Cityscapes & 2675 & 300 & 500 \\
Nighttime Driving & - & - & 50 \\
GTA-V & - & - & 25000 \\
CARLA & 16000 & 2000 & 2000 \\
Extreme parts. (GTA-V/City.) & - & - & 3750/100 \\
Foggy Cityscapes & 8025 & 900 & 1500 \\
Lum./contrasts Cityscapes & - & - & 3300 \\
Illuminants Cityscapes & - & - & 12000
\end{tabular}
\end{table}}

\section{Experiments and results}
\label{sec_experiments_results}

With the models and material exposed above, we perform different experiments, consisting of training and evaluating the networks presented in section \ref{sec_data_models}: the classical U-Net with no Divisive Normalization as baseline, and the modified U-Net with 4-DN layers. We take from \cite{HERNANDEZCAMARA202364} the models trained with the Cityscapes dataset and, following the same training procedure, we train more models with the synthetic CARLA dataset. We train each model ten times with the same ten different seeds as in \cite{HERNANDEZCAMARA202364}. Therefore, we have four models: with and without Divisive normalization trained with Cityscapes or CARLA.

\subsection{Data diversity}

First, we test and analyze the models with and without Divisive Normalization and trained with the real and synthetic images in different diverse datasets. This will allow us to analyze the effect of the Divisive Normalization trained with different data when copping with data diversity such as the nature of the data (real or synthetic), extreme conditions, day or night images or even the image resolution.

\subsubsection{Real vs synthetic}

First, we test how the results change between real and synthetic images. To do so, we test the models with the real images from the Cityscapes test and synthetic images from the CARLA test and GTA-V.

\begin{table}[h]
\caption{Mean IoU over ten trainings of models trained on real Cityscapes and synthetic CARLA and tested in: Cityscapes, CARLA and GTA-V. Improvements of the use of DN layers with regard to not using DN for each experiment in parenthesis.}
\label{data_divers_real_vs_sintetico}
\centering
\begin{tabular}{c|cc|cc}
\multicolumn{1}{c|}{\multirow{2}{*}{Dataset}} & \multicolumn{2}{c|}{Cityscapes trained} & \multicolumn{2}{c}{CARLA trained} \\ \cline{2-5} 
\multicolumn{1}{c|}{} & \multicolumn{1}{l|}{no-DN} & \multicolumn{1}{c|}{4-DN} & \multicolumn{1}{c|}{no-DN} & \multicolumn{1}{c}{4-DN} \\ \hline\hline
Cityscapes (real) & \multicolumn{1}{c|}{0.75} & 0.77 (2.7\%) & \multicolumn{1}{c|}{0.50} & 0.54 (8.0\%) \\ \hline
CARLA (synthetic) & \multicolumn{1}{c|}{0.51} & 0.54 (5.9\%) & \multicolumn{1}{c|}{0.90} & 0.91 (1.1\%) \\ \hline
GTA-V (synthetic) & \multicolumn{1}{c|}{0.55} & 0.61 (10.9\%) & \multicolumn{1}{c|}{0.62} & 0.65 (4.8\%) \\ \hline
\end{tabular}
\end{table}

Table \ref{data_divers_real_vs_sintetico} shows the mean IoU results over the ten training of the four models when evaluated in the different datasets. The first thing to notice is that the Divisive Normalization produces an improvement of the IoU results in all the scenarios. Moreover, it is interesting to note that each model gets better results in the test partition of the data it has been trained with. However, this is more extreme for the model trained with CARLA. This effect is due to the high difference between CARLA images and the other datasets as shown in Figure \ref{datasets_statistics}. Finally, the gain due to the Divisive Normalization is higher in the most different test data from the data the model has been trained with.

\subsubsection{Regular vs extreme images}
\label{results_global_extreme}

In \cite{HERNANDEZCAMARA202364} they found that the Divisive Normalization improvement increases with the fog level. It gives us the intuition that probably Divisive Normalization is even more important in extreme scenarios.

\begin{table}[h]
\caption{Mean IoU over ten trainings of models trained on Cityscapes good weather conditions and CARLA synthetic and tested in the different extreme partitions of Cityscapes and GTA-V. Improvements of the use of DN layers with regard to not using DN for each experiment in parenthesis. Improvements in particular extreme conditions higher than average improvement are marked in bold.}
\label{data_divers_global_vs_partitions}
\centering
\begin{tabular}{c|cc|cc}
\multicolumn{1}{c|}{\multirow{2}{*}{Dataset}} & \multicolumn{2}{c|}{Cityscapes trained} & \multicolumn{2}{c}{CARLA trained} \\ \cline{2-5} 
\multicolumn{1}{c|}{} & \multicolumn{1}{l|}{no-DN} & \multicolumn{1}{c|}{4-DN} & \multicolumn{1}{c|}{no-DN} & \multicolumn{1}{c}{4-DN} \\ \hline\hline

Cityscapes & \multicolumn{1}{c|}{0.75} & 0.77 (2.7\%) & \multicolumn{1}{c|}{0.50} & 0.54 (8.0\%) \\ \hline

\multicolumn{1}{c|}{City. (Low A)} & \multicolumn{1}{c|}{0.76} & 0.78 (\textbf{3.2\%}) & \multicolumn{1}{c|}{0.43} & 0.49 (\textbf{12.2\%}) \\ \hline

City. (High A) & \multicolumn{1}{c|}{0.74} & 0.77 (\textbf{3.1\%}) & \multicolumn{1}{c|}{0.54} & 0.56 (3.6\%) \\ \hline

City. (Low A ctr) & \multicolumn{1}{c|}{0.72} & 0.75 (\textbf{3.5\%}) & \multicolumn{1}{c|}{0.48} & 0.53 (\textbf{10.6\%}) \\ \hline

City. (High A ctr) & \multicolumn{1}{c|}{0.78} & 0.80 (2.6\%) & \multicolumn{1}{c|}{0.49} & 0.51 (5.0\%) \\ \hline

City. (Low clr ctr) & \multicolumn{1}{c|}{0.74} & 0.77 (\textbf{3.2\%}) & \multicolumn{1}{c|}{0.52} & 0.56 (\textbf{8.7\%}) \\ \hline

City. (High clr ctr) & \multicolumn{1}{c|}{0.74} & 0.76 (\textbf{3.1\%}) & \multicolumn{1}{c|}{0.46} & 0.52 (\textbf{11.5\%}) \\ \hline
\hline

GTA-V & \multicolumn{1}{c|}{0.55} & 0.61 (10.9\%) & \multicolumn{1}{c|}{0.62} & 0.65 (4.8\%) \\ \hline

\multicolumn{1}{c|}{GTA-V (Low A)} & \multicolumn{1}{c|}{0.46} & 0.52 (\textbf{13.0\%}) & \multicolumn{1}{c|}{0.55} & 0.58 (\textbf{5.5\%}) \\ \hline

GTA-V (High A) & \multicolumn{1}{c|}{0.61} & 0.67 (9.8\%) & \multicolumn{1}{c|}{0.68} & 0.69 (1.5\%) \\ \hline

GTA-V (Low A ctr) & \multicolumn{1}{c|}{0.57} & 0.63 (10.5\%) & \multicolumn{1}{c|}{0.65} & 0.67 (3.1\%) \\ \hline

GTA-V (High A ctr) & \multicolumn{1}{c|}{0.50} & 0.55 (10.0\%) & \multicolumn{1}{c|}{0.58} & 0.60 (3.4\%) \\ \hline

GTA-V (Low clr ctr) & \multicolumn{1}{c|}{0.50} & 0.57 (\textbf{14.0\%}) & \multicolumn{1}{c|}{0.60} & 0.63 (\textbf{5.0\%}) \\ \hline

GTA-V (High clr ctr) & \multicolumn{1}{c|}{0.54} & 0.59 (9.3\%) & \multicolumn{1}{c|}{0.60} & 0.62 (3.3\%) \\ \hline
\end{tabular}
\end{table}

Table \ref{data_divers_global_vs_partitions} shows the results of the models when evaluated in the different extreme partitions of Cityscapes and GTA-V. Again, the Divisive Normalization layers always help the models to get better results. Moreover, the DN improvements are even higher than in the complete dataset for almost all the extreme images in the Cityscapes dataset and especially for the models trained in the Cityscapes real images. Note that for the synthetic GTA-V extreme partitions, the DN gain is higher than the average in the low luminance (dark images) and in the low color contrast images. This raises the question of how the models with Divisive Normalization would perform in very dark images, i.e. extremely low-luminance images such as night images.

\subsubsection{Day vs night}

To answer our previous question, we test our models in a dataset made of real nighttime images. These images have extremely low luminance and we expect a high gain when using the Disive Normalization.

\begin{table}[h]
\caption{Mean IoU over ten trainings of models trained on Cityscapes good weather conditions and CARLA synthetic and tested in: Cityscapes, Nighttime Driving, CARLA and GTA-V. Improvements of the use of DN layers with regard to not using DN for each experiment in parenthesis.}
\label{data_divers_day_vs_night}
\centering
\begin{tabular}{c|cc|cc}
\multicolumn{1}{c|}{\multirow{2}{*}{Dataset}} & \multicolumn{2}{c|}{Cityscapes trained} & \multicolumn{2}{c}{CARLA trained} \\ \cline{2-5} 
\multicolumn{1}{c|}{} & \multicolumn{1}{l|}{no-DN} & \multicolumn{1}{c|}{4-DN} & \multicolumn{1}{c|}{no-DN} & \multicolumn{1}{c}{4-DN} \\ \hline\hline
Cityscapes (real-day) & \multicolumn{1}{c|}{0.75} & 0.77 (2.7\%) & \multicolumn{1}{c|}{0.50} & 0.54 (8.0\%) \\ \hline
\multicolumn{1}{c|}{Nighttime (real-night)} & \multicolumn{1}{c|}{0.24} & \multicolumn{1}{c|}{0.29 (20.8\%)} & \multicolumn{1}{c|}{0.31} & \multicolumn{1}{c}{0.33 (6.5\%)} \\ \hline
CARLA (synthetic) & \multicolumn{1}{c|}{0.51} & 0.54 (5.9\%) & \multicolumn{1}{c|}{0.90} & 0.91 (1.1\%) \\ \hline
GTA-V (synthetic) & \multicolumn{1}{c|}{0.55} & 0.61 (10.9\%) & \multicolumn{1}{c|}{0.62} & 0.65 (4.8\%) \\ \hline
\end{tabular}
\end{table}

Table \ref{data_divers_day_vs_night} shows the results of the models when evaluated in the Nighttime Driving dataset. As expected due to the extremely low-luminance, the worst results for all the models happen in the night images. However, in this dataset is where the models trained in real data get a higher increase due to the DN layers, showing its importance. Models trained in CARLA do not get the higher increase due to the DN layers in the night images because the CARLA dataset has some night images and therefore these models have seen low luminance images during their training. In fact, for these models, the DN higher effect happens when facing real daytime images, because as shown in Figure \ref{datasets_statistics} it has really different statistics from real daytime images.

\subsubsection{High vs low resolution}

To test how the image resolution affects the results, we retrain the models with the Cityscapes data but maintain the images at their full resolution, without any resizing. Therefore, we train the models exactly in the same way except for the image resolution.
Due to the high computational cost of training models with this high image resolution, we only train a single model with the Cityscapes dataset.

Table \ref{data_divers_low_vs_full_res} shows that training with low or high-resolution images does not give too much difference in the overall IoU results or in the effect of the Divisive Normalization.

\begin{table}[h]
\caption{Mean IoU of low and full resolution models trained on Cityscapes in good weather conditions and tested in Cityscapes. Improvements of the use of DN layers with regard to not using DN for each experiment in parenthesis.}
\label{data_divers_low_vs_full_res}
\centering
\begin{tabular}{c|cc|cc}
\multicolumn{1}{c|}{\multirow{2}{*}{Dataset}} & \multicolumn{2}{c|}{Low resolution trained} & \multicolumn{2}{c}{Full resolution trained} \\ \cline{2-5} 
\multicolumn{1}{c|}{} & \multicolumn{1}{l|}{no-DN} & \multicolumn{1}{c|}{4-DN} & \multicolumn{1}{c|}{no-DN} & \multicolumn{1}{c}{4-DN} \\ \hline\hline
Cityscapes & \multicolumn{1}{c|}{0.75} & 0.77 (2.7\%) & \multicolumn{1}{c|}{0.73} & 0.76 (4.1\%) \\ \hline
\end{tabular}
\end{table}

\subsection{Controled changes of environment}

Previous section results are really good and show that 1) the Divisive Normalization always helps the models to cope with the data variability and improve their results and 2) the use of the Divisive Normalization is even more important in extreme scenarios, such as extremely low-luminance images as night images, and also low and high chromatic contrast images. However, the images we used to understand the effect of Divisive Normalization come from different datasets with different statistics. To analyze the effect of the DN in the models in a completely controlled scenario, we need to maintain always the same base images. In that way, if we do some variation always to the same images, the only variations in the results will come from facing the introduced changes. Therefore, in this section, we select a real dataset, Cityscapes and we will perform some experiments with variations of it.

\subsubsection{Fog change}

First, we use the Foggy Cityscapes dataset. It consists of the same Cityscapes images with synthetic fog added to simulate three different fog severities. Using these three modified datasets we can test the effect of the Divisive Normalization when facing images with progressively more fog, i.e. progressively more luminance and less contrast as shown in Figure \ref{datasets_statistics}. In addition to the two models trained on the Cityscapes and CARLA datasets, we develop a new model trained in a combination of Cityscapes and Foggy Cityspaces train images (of the three severities), so that we can compare the results between a model trained in completely synthetic images (CARLA model), real clean daytime images (Cityscapes model) and a model trained in real images but with more variability (Cityscapes + Foggy Cityscapes model).

\begin{table}[h]
\caption{Mean IoU over ten models trained on Cityscapes, Cityscapes + Foggy Cityscapes and CARLA and tested in Cityscapes, the different Foggy Cityscapes severities, Nighttime Driving, CARLA and GTA-V. Improvements of the use of DN layers with regard to not using DN for each experiment in parenthesis. Results marked with * are from \cite{HERNANDEZCAMARA202364}.}
\label{controled_changes_fog}
\centering
\begin{tabular}{|c|cc|cc|cc|}
\hline
\multirow{2}{*}{Dataset} & \multicolumn{2}{c|}{City. train} & \multicolumn{2}{c|}{City. + Foggy train} & \multicolumn{2}{c|}{CARLA train} \\ \cline{2-7} 

 & \multicolumn{1}{c|}{no-DN} & 4-DN & \multicolumn{1}{c|}{no-DN} & 4-DN & \multicolumn{1}{c|}{no-DN} & 4-DN \\ \hline
 
Cityscapes & \multicolumn{1}{c|}{*0.75} & *0.77 (2.7\%) & \multicolumn{1}{c|}{0.78} & 0.79 (1.2\%) & \multicolumn{1}{c|}{0.50} & 0.54 (8.0\%) \\ \hline

Foggy (low) & \multicolumn{1}{c|}{*0.65} & *0.70 (7.7\%) & \multicolumn{1}{c|}{0.78} & 0.79 (1.2\%) & \multicolumn{1}{c|}{0.46} & 0.52 (13.0\%) \\ \hline

Foggy (middle) & \multicolumn{1}{c|}{*0.54} & *0.62 (14.8\%) & \multicolumn{1}{c|}{0.78} & 0.79 (1.2\%) & \multicolumn{1}{c|}{0.44} & 0.50 (13.6\%) \\ \hline

Foggy (high) & \multicolumn{1}{c|}{*0.40} & *0.48 (22.5\%) & \multicolumn{1}{c|}{0.77} & 0.78 (1.8\%) & \multicolumn{1}{c|}{0.40} & 0.46 (15.0\%) \\ \hline

Nighttime & \multicolumn{1}{c|}{0.24} & 0.29 (20.8\%) & \multicolumn{1}{c|}{0.34} & 0.36 (5.9\%) & \multicolumn{1}{c|}{0.31} & 0.33 (6.5\%) \\ \hline

CARLA & \multicolumn{1}{c|}{0.51} & 0.54 (5.9\%) & \multicolumn{1}{c|}{0.58} & 0.67 (16.7\%) & \multicolumn{1}{c|}{0.90} & 0.91 (1.1\%) \\ \hline

GTA-V & \multicolumn{1}{c|}{0.55} & 0.61 (10.9\%) & \multicolumn{1}{c|}{0.65} & 0.68 (3.6\%) & \multicolumn{1}{c|}{0.62} & 0.65 (4.8\%) \\ \hline
\end{tabular}
\end{table}

Table \ref{controled_changes_fog} shows the results of the three models. \cite{HERNANDEZCAMARA202364} found that the improvements due to the Divisive Normalization increase with fog level for a model trained in Cityscapes. We found it also happens with a model trained with synthetic images. As expected, the model trained with a combination of Cityscapes and Foggy Cityscapes gets much more constant results because it has seen foggy images during its training. However, the DN layers still have a positive impact and the gain gets higher in high fog level. Interestingly, gains are more singnificant when dealing with very different data as the synthetic scenes.

\subsubsection{Luminance and contrasts changes}
\label{sec_sum_and_ctr}

Our previous results highlight that the Divisive Normalization is even more important in very dark images and as the fog level increases. Motivated by these results and to get a deeper analysis of the Divisive Normalization gains, we have to control the exact luminance and contrast of the images. We test the models with 3D (luminance, achromatic contrast and chromatic contrast) modified tensor images.

\begin{figure}
\centering
  \includegraphics[width=0.9\linewidth]{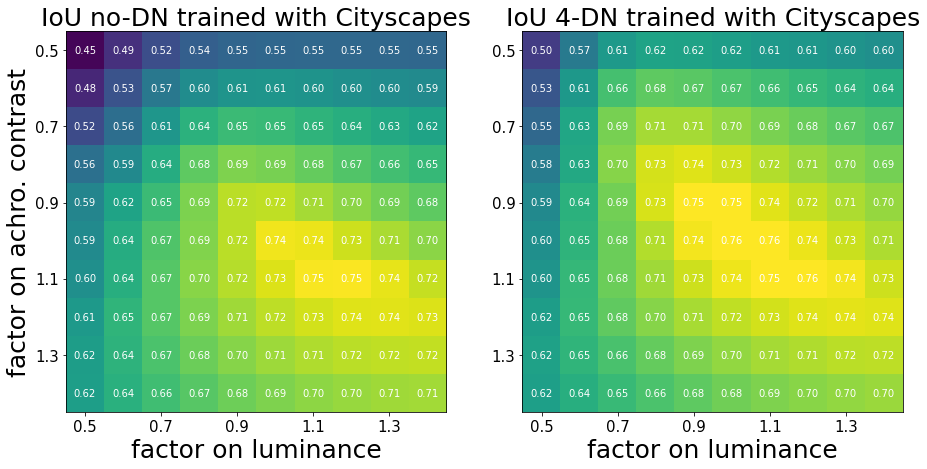}\\
  \caption{IoU of the no-DN (left) and 4-DN (right) model trained with Cityscapes images depending on the luminance and achromatic contrast of the input images. In this example, the chromatic contrast of the images has not been changed.} 
  \label{cityscapes_slice_fixed_chromatic_contrast}
\end{figure}

Figure \ref{cityscapes_slice_fixed_chromatic_contrast} shows how the IoU results of the no-DN and 4-DN models trained with the Cityscapes images depend on the luminance and achromatic contrast of the input images, with a fixed chromatic contrast equal to its original value, such as in the green slice in figure \ref{tensor_example}. As expected, the lowest IoU's are obtained in the low luminance and low achromatic contrast conditions. The models that implement the Divisive Normalization layers not only get higher IoU values but the region where they obtain higher values is bigger than for the no-DN models.

\begin{figure}[h]
\begin{center}
\includegraphics[width=1.0\linewidth]{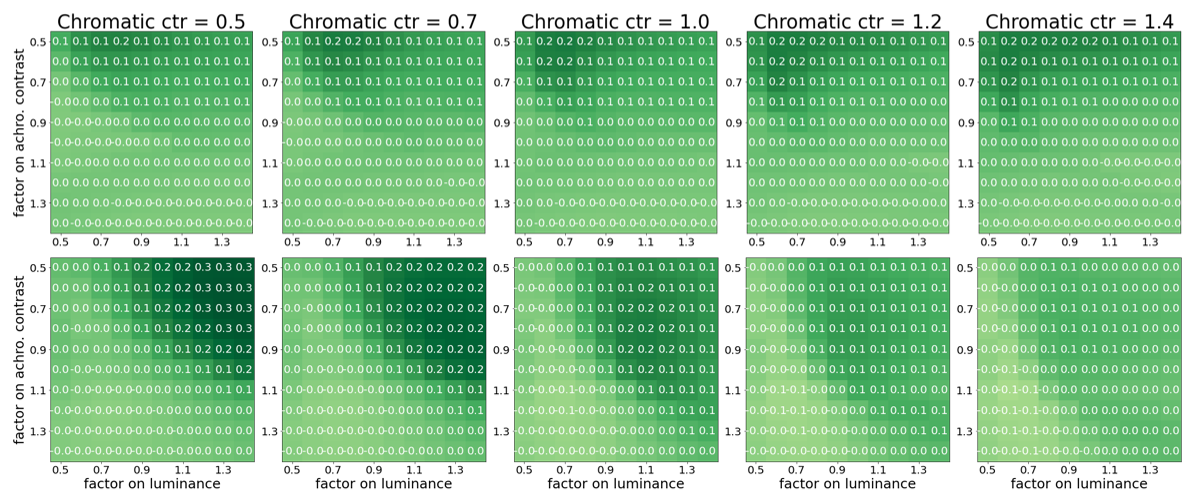}\\
  \caption{IoU gain of the 4-DN models with regard to the no-DN models trained with different data depending on the luminance, achromatic contrast and chromatic contrast of the input images. We show five different slides for five different chromatic contrasts. \emph{Top:} models trained in Cityscapes. \emph{Bottom:} models trained in CARLA.}
  \label{DN_gains_tensor_img}
    \end{center}
\end{figure}

\begin{figure}
\centering
  \includegraphics[width=0.9\linewidth]{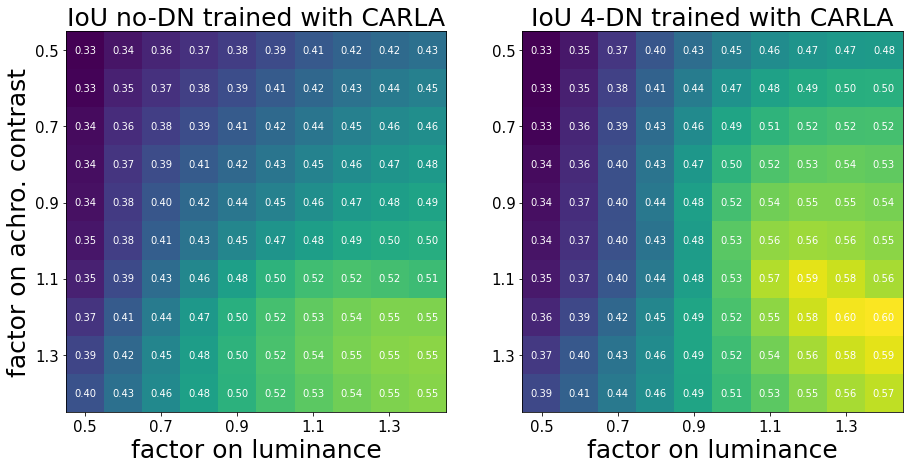}\\
  \caption{IoU of the no-DN (left) and 4-DN (right) model trained with CARLA images depending on the luminance and achromatic contrast of the input images. In this example, the chromatic contrast of the images has not been changed.} \label{carla_slice_fixed_chromatic_contrast}
\end{figure}

It is easier to visualize the benefit of the Divisive Normalization using the gain of the 4-DN models with regard to the no-DN models as shown in figure \ref{DN_gains_tensor_img} for different values of chromatic contrast and for the models trained with the different data. The first row shows the gains for the model trained with the Cityscapes data and the second row shows the gains for the models trained with the CARLA synthetic data. We can obtain different conclusions from this figure. First, we see that using the Divisive Normalization in the models always helps to achieve better results.  Second, the models trained in Cityscapes get the maximum gain at low contrast and luminances, implying that it is especially important for night and foggy conditions. However, what happens when we focus on the models trained with the CARLA synthetic images? We can see that although there is still a high gain region of the Divisive Normalization, it is not located at low luminances and contrasts as before but the greatest improvements happen at high luminances. To understand what is happening, in figure \ref{carla_slice_fixed_chromatic_contrast} we plot the IoU values of the no-DN and 4-DN models when fixing the chromatic contrast to its original value, as we did in figure \ref{cityscapes_slice_fixed_chromatic_contrast}. We get that the maximum IoU the models get is highly displaced from the 1-1 original conditions. It implies that the synthetic CARLA images have higher luminances and contrasts than real Cityscapes images and that is why the models get their highest IoU at higher luminances and contrasts than the real images. Also, it explains why the gain region is displaced from the top-left corner in the bottom panel of figure \ref{DN_gains_tensor_img}. The gain happens at lower luminances and contrasts than the original image conditions, which in the CARLA dataset are displaced to higher luminances and contrasts and so the gain region is also displaced.

\subsubsection{Illuminant changes}

\begin{figure}[h]
\centering
  \includegraphics[width=\linewidth]{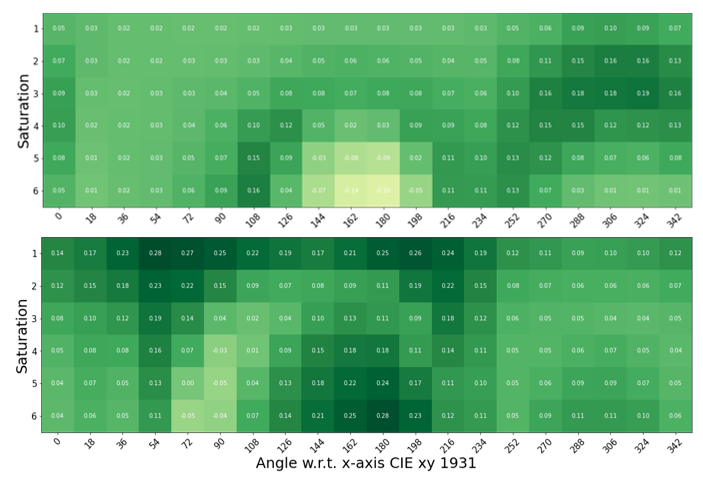}\\
  \caption{IoU gain of the 4-DN models with regard to the no-DN models trained with the different datasets depending on illuminant hue and saturation. \emph{Top:} models trained in Cityscapes. \emph{Bottom:} models trained in CARLA.} \label{illuminant_cityscapes_results}
\end{figure}

We also test our models with the images that we modified lighting them with different illuminants of controlled dominant wavelength and saturation. Figure \ref{illuminant_cityscapes_results} shows directly the gain of using 4-DN models with regard to the no-DN models when they face the different illuminant images for the models trained with the different data. We obtained that including the Divisive Normalization in the models trained with Cityscapes images (first row) improves the segmentation results in almost all the illuminant-intensity space, except for the model trained with Cityscapes data in just a small region of very high saturations, which are almost impossible images in real life.

In the CARLA training scenario (second row), we see that the gain region of the Divisive Normalization models trained with CARLA does not follow the same trend as when the models are trained with real images but again they show higher gains due to the Divisive Normalization, especially in the low saturation.

\section{Analysis: invariance and nonlinearities}
\label{analysis_section}

First, why do the models that include Divisive Normalization always achieve better segmentation performance?
We address this question through quantitative measures of the invariance of the prediction space under changes in the environment. These measures below suggest that segmentation is better because the representation of the networks using Divisive Normalization is more invariant. 

Second, what is the qualitative effect of Divisive Normalization 
in the responses of neurons at progressively deeper layers?
In the models trained for image segmentation, we use a specific stimulation of neurons after Divisive Normalization (in each layer injecting certain activity in the neighbour neurons) to check the response to the energy of the corresponding feature.
By doing this, we show that the responses after DNs are nonlinear, and the gain of the nonlinearity reduces with the activity of the neighbor neurons (as expected from the analytical expression of DN).
This specific behaviour is good for equalizing the response of all neurons in a feature map~\cite{Malo10}, and this is good for obtaining an invariant representation. 
In a model trained for segmentation, we show that this input-dependent inhibition increases for deeper layers.  

\subsection{Quantitative measures of the invariance}

\begin{figure*}
\centering
  \includegraphics[width=0.8\textwidth]{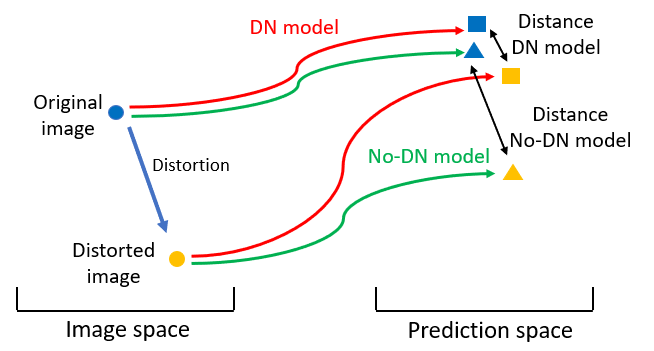}\\
  \caption{\textbf{Quantification of invariance.} If a model presents invariance to some distortion, it means that images that are further away in their original space due to that distortion are transformed to close points in the model prediction domain. 
  If the DN induces invariance, the distance between the predictions of the DN model should be smaller than the distance between the predictions of the no-DN model.}\label{invariance}
\end{figure*}

Figure~\ref{invariance} describes the hypothesis we have on the effect of Divisive Normalization on the invariance of the signal representation when the input undergoes changes in the environment. 
Given the qualitative behavior expected from the literature on biological adaptation~\cite{Carandini2012NormalizationAA,Malo10}, confirmed in the next section for networks trained for segmentation, one expects that samples that have been put away in the input domain due to changes in the environment will remain close in the inner representation of systems augmented with Divisive Normalization. 
In contrast, conventional systems would keep these samples separated in the inner representation. In other words, the inner representation of the nets with Divisive Normalization will be more invariant under environmental changes than the representation in conventional nets (with no DN). 
This invariance can be quantitatively assessed by measuring the overlap between the responses of the networks to original and distorted inputs. This is what we do in the experiment shown here.

To do so, we compute the prediction of the no-DN and 4-DN models for the original and the modified images (under fog, luminance, contrast and illuminant changes). Then, we calculate the IoU metric between the prediction of the original image and the prediction of the modified image.
This IoU-overlap measure is a measure of the \emph{lenght} of the black arrows in the inner domain shown in Figure~\ref{invariance}: bigger IoU implies more overlap and smaller distance.

\begin{table}
\centering
\caption{Iou between the prediction on the original image and the prediction on the modified image for some of the modifications, comparing the results of the models with and without Divisive Normalization layers.}
\label{invariances}
\begin{tabular}{c|c|c}  
Dataset & no-DN IoU($pred_{ori}$, $pred_{mod}$) & 4-DN IoU($pred_{ori}$, $pred_{mod}$) \\  \hline\hline
Low fog & 0.766 & 0.826 \\
Middle fog & 0.621 & 0.714 \\
High fog & 0.478 & 0.609 \\ \hline
Achrom ctr = 0.6 & 0.649 & 0.786 \\
Achrom ctr = 1.4 & 0.793 & 0.832 \\
Luminance = 0.6 & 0.739 & 0.781 \\
Luminance = 1.4 & 0.848 & 0.886 \\ 
Chrom ctr = 0.6 & 0.842 & 0.844 \\
Chrom ctr = 1.4 & 0.848 & 0.887 \\ \hline
Angle 0 & 0.464 & 0.800 \\
Angle 72 & 0.751 & 0.826 \\
Angle 162 & 0.517 & 0.562 \\
Angle 252 & 0.657 & 0.772 \\
Angle 343 & 0.365 & 0.753 \\
\end{tabular}
\end{table}

Table \ref{invariances} shows the IoU results between the predictions over the original images and the predictions over the modified images both for the no-DN and 4-DN models. It shows results for all the modifications we tested: fog, change of luminance, achromatic and chromatic contrasts and illuminants. We got that for all the modifications, the IoUs of the 4-DN model are always higher than the IoUs of the no-DN model. This implies that the predictions of the model that implements the Divisive Normalization change less under all the image changes than the ones of the no-DN model.

\subsection{Where does the invariance come from?: adaptive nonlinearities}

\begin{figure*}[t]
\centering
  \includegraphics[width=0.75\textwidth]{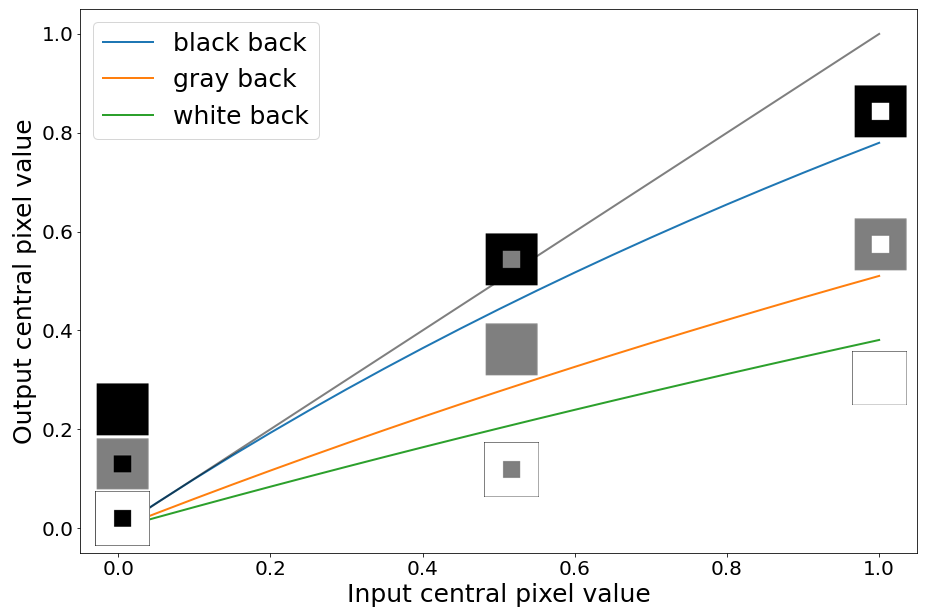}\\
  \caption{\textbf{Checking the nonlinearity of responses.}
  Effect of the first Divisive Normalization model layers on some example images that have different central pixel values and different surrounding values. It shows how the central pixel normalized value depends not only on its value but also on the neighbour pixels.}
  \label{DN_first_analysis}
\end{figure*}

\begin{figure*}[b]
\centering
  \includegraphics[width=1.0\textwidth]{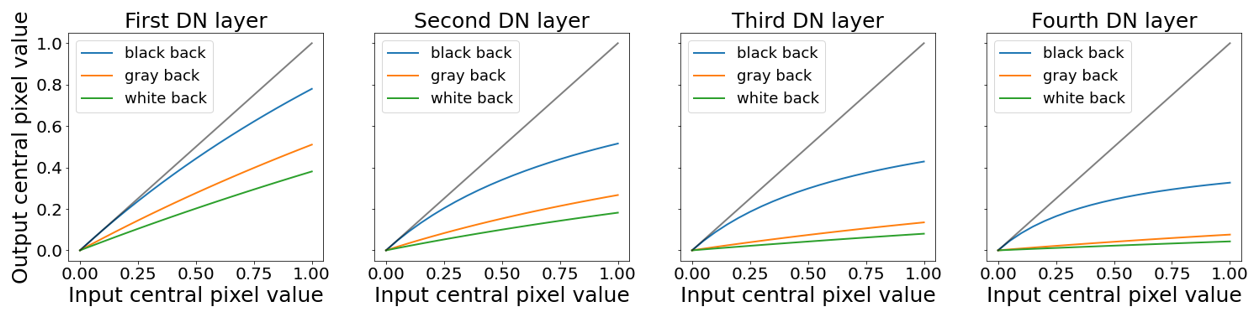}\\
  \caption{\textbf{Nonlinearities at different depths}. Effect of the four Divisive Normalization model layers depending on the pixel value and its surround, showing the adaptation of the normalization.}
  \label{DN_layers_analysis}
\end{figure*}

We argue that invariance comes from the adaptation of the responses of each neuron to the responses of the neighbor neurons that is enforced by the local normalization. 
In order to illustrate this intuition, we visualize the response of 
the response of neurons of the four Divisive Normalization layers obtained in a model trained for segmentation.
In particular, we build $3 \times 3$ (same size as the $\gamma$ kernel in the Divisive Normalization denominator) neighborhoods where the central pixel value changes between 0 and 1 (or the minimum and maximum of the corresponding input feature). We try different surrounding pixel values such as $0$, $0.5$, and $1.0$ (or equivalent fractions of the dynamic range of the input): implying a black, grey, or white background. We pass these images through the four Divisive Normalization layers and register how the central pixel value has been transformed depending on its own value and the value of the surround. Figure \ref{DN_first_analysis} shows some of the generated images and the effect of the first divisive normalization layer depending on the central pixel and surrounding values. Figure \ref{DN_layers_analysis} shows the effect for the four Divisive Normalization layers.

We obtain that inputs where the background is very active are strongly inhibited with regard to inputs with inactive backgrounds: the central pixel is more inhibited as the background changes from completely black (surrounding pixel values of 0 and then only the central value is considered) to white (surrounding pixel values of 1, which are the higher values they can have). Looking at the different Divisive Normalization layers we observe this trend in the four layers. Still, it is interesting to note that the non-linearity (departure from the reference black line) increases its effect with the layer depth because the ratio between $\beta$ over $\gamma$ value (which controls the linear/non-linear behaviour) found in our training is reduced with the layer depth.

The two properties of these curves tend to \emph{equalize} the responses: (1)~the saturating shape of the blue curves amplifies low values and moderates high values, and (2)~the presence of high values (in the neighbours) moderates the responses (attenuated orange and green curves), while the presence of low values amplifies the responses.  
The first property transforms the input responses (with eventually disparate range) into output responses within a more compact and stable range performing a sort of univariate histogram equalization. And the second property makes the range of the outputs more equal along (spatial/feature) regions where the inputs had different ranges.
The combined effect of these properties makes the range of the output responses kind of invariant to changes in the range of the input, probably leading to the invariance measured in the previous section. 

These nonlinearities had been described before in vision science to explain adaptation to luminance, color~\cite{Abrams07} and contrast~\cite{Watson97}, and their effect on multivariate equalization has been extensively illustrated too~\cite{Schwartz01,Malo10,Martinez18,Malo20}. The interesting novelty here is that this behavior, convenient for invariance, easily emerged in networks trained for segmentation because we used the layer with the appropriate analytical expression (or the appropriate capacity).

\section{Conclusion}
\label{sec_conclusion}

\begin{figure}[h]
     \begin{subfigure}[b]{0.45\textwidth}
         \centering
         \includegraphics[width=\textwidth]{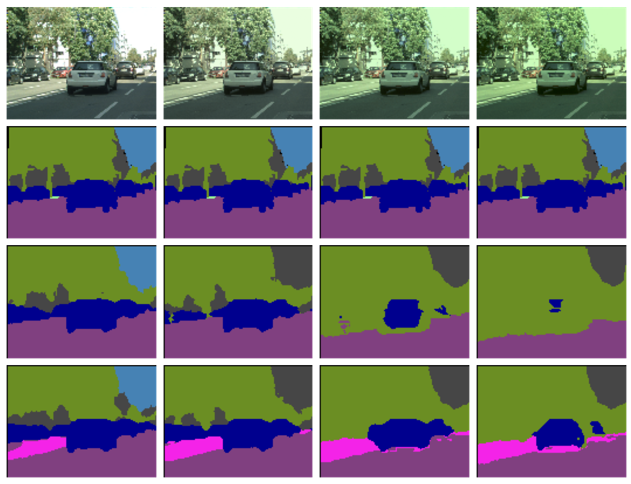}
         \caption{Illuminant changes.}
         \label{invariance_illuminant}
     \end{subfigure}
     \hfill
     \begin{subfigure}[b]{0.45\textwidth}
         \centering
         \includegraphics[width=\textwidth]{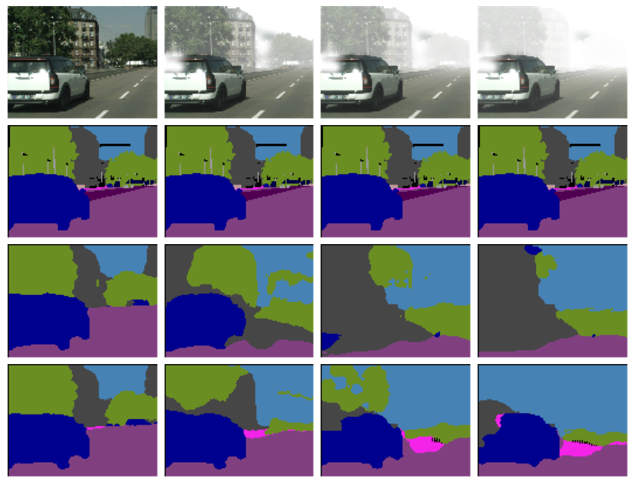}
         \caption{Fog changes.}
         \label{invariance_fog}
     \end{subfigure}

     \vfill

     \begin{subfigure}[b]{0.45\textwidth}
         \centering
         \includegraphics[width=\textwidth]{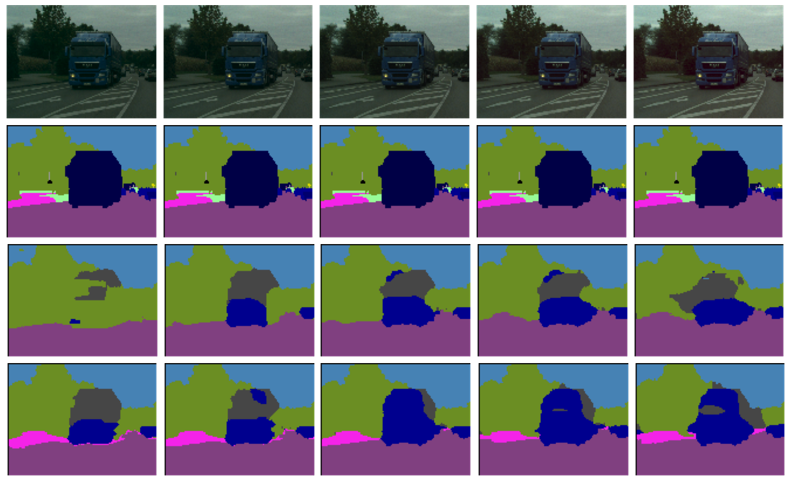}
         \caption{Luminance changes.}
         \label{invariance_luminance}
     \end{subfigure}
     \hfill
     \begin{subfigure}[b]{0.45\textwidth}
         \centering
         \includegraphics[width=\textwidth]{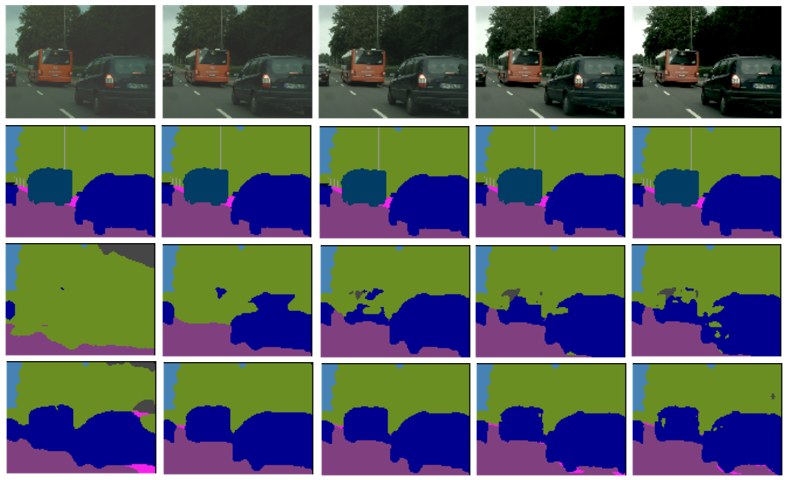}
         \caption{Achromatic contrast changes.}
         \label{invariance_contrast}
     \end{subfigure}

\caption{\textbf{Summary: dealing with diversity}. Example of the models' predictions under the different controlled changes. In each panel, the first row shows the input images; the second row shows the segmentation ground truth; and the third and fourth rows show the predictions of the no-DN and 4-DN models respectively. Panel \ref{invariance_illuminant} and \ref{invariance_fog} show the predictions when the saturation of an illuminant and the fog level increases (left to right). Panel \ref{invariance_luminance} and \ref{invariance_contrast} show the predictions when the image mean luminance and achromatic contrast are reduced (from the central image to the left) or increased (from the central image to the right).}
\label{figura_final}
\end{figure}

We have shown that introducing the Divisive Normalization layer in segmentation models helps them to achieve better results in all the tested scenarios, not only in a range of natural and synthetic databases, but also when we systematically extended the changes over five extra visual dimensions: luminance, achromatic contrast, chromatic contrast, spectral illumination (i.e. hue, and saturation). We also found that the use of Divisive Normalization becomes more important in extreme scenarios, such as low luminances (night images) or low contrast (high fog). Moreover, we discovered that these gains come from making the models with Divisive Normalization more invariant under different environmental changes and therefore help them to achieve better results in all the tested conditions. These effects are visually summarized in figure \ref{figura_final} which shows how the model with Divisive Normalization gets similar segmentation results when environment changes are applied to the image, while the segmentation produced by the classical U-Net is clearly affected by these changes.

More detailed, panel \ref{invariance_illuminant} shows predictions when changing the illuminant by increasing the saturation. When the saturation is high, the car (segmented in blue) disappears from the no-DN prediction. Panel \ref{invariance_fog} shows results for different levels of fog, this variation also affects the no-DN model by stopping to detect a car even for the middle fog level. Panels \ref{invariance_luminance} and \ref{invariance_contrast} show how the prediction changes when the luminance and achromatic contrast are increased or reduced. In agreement with results obtained in sec \ref{sec_sum_and_ctr}, there are higher gains due to the Divisive Normalization in the low luminances and low contrast. When the luminance decreases the no-DN model does not detect the truck while the 4-DN model still detects part of it. The same happens when the contrast gets reduced, the no-DN model completely mismatches the car, labelling it as vegetation.

To conclude, our results show that including the Divisive Normalization in the segmentation algorithm makes them invariant under many image changes, which is helpful for example for autonomous driving.

\section*{Acknowledgments}

This work was supported in part by MICIIN/FEDER/UE under Grant PID2023-152133NB-I00; in part by Spanish MIU under Grant FPU21/02256; and in part by BBVA Foundations of Science program: Maths, Stats, Comp. Sci. and AI (VIS4NN). The authors gratefully acknowledge the computer resources at Artemisa and the technical support provided by the European Union through the 2014–2020 ERDF Operative Programme of Comunitat Valenciana, project IDIFEDER/2018/048.



\bibliographystyle{elsarticle-num} 
\bibliography{thebibliography}

@inproceedings{Cordts2016Cityscapes,
title = "The Cityscapes Dataset for Semantic Urban Scene Understanding",
author = "Cordts, M. and others",
booktitle = "2016 CVPR",
year = "2016"}

@article{Orallo23,
volume = {380},
number = {6641},
author = {R Burnell and others},
title = {Rethink reporting of evaluation results in {AI}},
year = {2023},
journal = {Science},
pages = {136--138}}

@misc{zhou2023,
title={Predictable Artificial Intelligence}, 
author={Lexin Zhou and others},
year={2023},
eprint={2310.06167},
archivePrefix={arXiv}}

@article{Colorlab02,
author = {Malo, J. and  Luque,  M.J.},
journal = {Univ. Valencia.  http://isp.uv.es/code/visioncolor/colorlab.html},
title = {{ColorLab: A Matlab Toolbox for Color Science and Calibrated Color Image Processing}},
year = {2002},
url = {http://isp.uv.es/code/visioncolor/colorlab.html}}

@BOOK{Stiles00,
AUTHOR = "Wyszecki, G and Stiles, WS.",
TITLE  = "Color Science: Concepts and Methods, Quantitative Data and Formulae",
PUBLISHER = "John Wiley \& Sons",
YEAR = 2000,
ADDRESS = "New Jersey"}

@ARTICLE{Martinez19,
AUTHOR={Martinez, M.  and Bertalmío, M. and Malo, J.},
TITLE={In Praise of Artifice Reloaded: Caution With Natural Image Databases in Modeling Vision},
JOURNAL={Frontiers in Neuroscience},
VOLUME={13},
YEAR={2019}}

@inproceedings{fairchilddatabasehdr,
  title={The {HDR} photographic survey},
  author={Fairchild, Mark D},
  booktitle={Color and imaging conference},
  volume={15},
  pages={233--238},
  year={2007}}

@article{deeb2018interreflections,
title={Interreflections in computer vision: a survey and an introduction to spectral infinite-bounce model},
author={Deeb, R. and Muselet, D. and others},
journal={Journal of Mathematical Imaging and Vision},
volume={60},
pages={661--680},
year={2018}}

@article{laparra2012nonlinearities,
title={Nonlinearities and adaptation of color vision from sequential principal curves analysis},
author={Laparra, V. and Jim{\'e}nez, S. and others},
journal={Neural Computation},
volume={24},
number={10},
pages={2751--2788},
year={2012}}

@misc{IPL_Carla_dataset,
title={IPL-CARLA-dataset },
howpublished={\url{https://huggingface.co/datasets/isp-uv-es/IPL-CARLA-dataset}},
note = {Accessed: 2024-07-10}}

@misc{IPL_City_Illuminants,
title={IPL-Cityscapes-Illuminants},
howpublished = {\url{https://huggingface.co/datasets/isp-uv-es/IPL-Cityscapes-Illuminants}},
note = {Accessed: 2024-07-10}}

@misc{IPL_City_LuminanceContrast,
title={
IPL-Cityscapes-LuminanceContrasts},
howpublished = {\url{https://huggingface.co/datasets/isp-uv-es/IPL-Cityscapes-LuminanceContrasts}},
note = {Accessed: 2024-07-10}}

@article{fog_cityscapes,
author = "Sakaridis, C. and others",
title = "Semantic Foggy Scene Understanding with Synthetic Data",
journal = "Int. J. Comput. Vis.",
year = "2018",
volume = "126",
number = "9",
pages = "973-992"}

@Inbook{Camps11,
author="Camps-Valls, G. and Tuia, D. and others",
title="The Statistics of Remote Sensing Images",
bookTitle="Remote Sensing Image Processing",
year="2012",
publisher="Springer International Publishing",
pages="26--47",
isbn="978-3-031-02247-0"}

@book{Cavanagh18,
  title={The visual world of shadows},
  author={Casati, R. and Cavanagh, P.},
  year={2023},
  publisher={MIT Press}}

@article{Jimenez13,
  title={The role of spatial information in disentangling the irradiance--reflectance--transmittance ambiguity},
  author={Jim{\'e}nez, S. and Malo, J.},
  journal={IEEE Transactions on geoscience and remote sensing},
  volume={52},
  number={8},
  pages={4881--4894},
  year={2013}}

@inproceedings{daytime_2_nighttime, 
  author = {Dai, Dengxin and {Van Gool}, Luc}, 
  booktitle = {IEEE International Conference on Intelligent Transportation Systems}, 
  title = {Dark Model Adaptation: Semantic Image Segmentation from Daytime to Nighttime}, 
  year = {2018}}

@inproceedings{Carla_dataset,
  title = {{CARLA}: {An} Open Urban Driving Simulator},
  author = {Alexey Dosovitskiy and German Ros and others},
  booktitle = {Proceedings of the 1st Annual Conference on Robot Learning},
  pages = {1--16},
  year = {2017}}

@inproceedings{gta,
  title={Playing for data: Ground truth from computer games},
  author={Richter, Stephan R and Vineet, Vibhav and Roth, Stefan and Koltun, Vladlen},
  booktitle={European Conference on Computer Vision},
  pages={102--118},
  year={2016}}

@book{fairchild2013color,
title={Color appearance models},
author={Fairchild, M.D.},
year={2013},
publisher={John Wiley \& Sons}}

@inproceedings{saleh2018effective,
title={Effective use of synthetic data for urban scene semantic segmentation},
author={Saleh, F.S. and Aliakbarian, M.S. and others},
booktitle={2018 ECCV},
pages={84--100},
year={2018}}

@inproceedings{ros2016synthia,
title={The synthia dataset: A large collection of synthetic images for semantic segmentation of urban scenes},
author={Ros, G. and Sellart, L. and others},
booktitle={2016 CVPR},
pages={3234--3243},
year={2016}}

@misc{Tesla,
title={Tesla’s Autopilot and Full Self-Driving linked to hundreds of crashes, dozens of deaths},
author={A.J. Hawkins},
year=2024,
note={\url{https://www.theverge.com/2024/4/26/24141361/tesla-autopilot-fsd-nhtsa-investigation-report-crash-death} [Accessed: 08/07/2024]}}

@article{HERNANDEZCAMARA202364,
title = {Neural networks with divisive normalization for image segmentation},
journal = {Pattern Recognition Letters},
volume = {173},
pages = {64-71},
year = {2023},
author = {Pablo Hernández-Cámara and Jorge Vila-Tomás and Valero Laparra and Jesús Malo}}

@article{Carandini2012NormalizationAA,
title = "Normalization as a canonical neural computation",
author = "Carandini, Matteo and Heeger, David J",
journal = "Nat. Rev. Neurosci.",
year = "2012",
number = 1,
volume = "13",
pages = "51-62"}

@article{Malo20,
title={Spatio-chromatic information available from different neural layers via Gaussianization},
author={Malo, J.},
journal={J. Math. Neurosci.},
volume={10},
number={1},
pages={1-40},
year={2020}}

@article{Malo10,
title={Psychophysically tuned divisive normalization approximately factorizes the {PDF} of natural images},
author={Malo, Jes{\'u}s and Laparra, Valero},
journal={Neural Computation},
volume={22},
number={12},
pages={3179-3206},
year={2010},}

@article{Martinez18,
author = {Martinez-Garcia, M. AND Cyriac, P. AND Batard, T. AND Bertalmío, M. AND Malo, J.},
journal = {PLOS ONE},
publisher = {Public Library of Science},
title = {Derivatives and inverse of cascaded linear+nonlinear neural models},
year = {2018},
volume = {13},
pages = {1-49}}

@article{Abrams07,
author = {Abrams, Alicia B. and Hillis, James M. and Brainard, David H.},
title = "{The Relation Between Color Discrimination and Color Constancy: When Is Optimal Adaptation Task Dependent?}",
journal = {Neural Computation},
volume = {19},
number = {10},
pages = {2610-2637},
year = {2007}}

@article{Watson97,
author = {Andrew B. Watson and Joshua A. Solomon},
journal = {J. Opt. Soc. Am. A},
number = {9},
pages = {2379--2391},
title = {Model of visual contrast gain control and pattern masking},
volume = {14},
year = {1997}}

@article{Schwartz01,
title = "Natural signal statistics and sensory gain control",
author = "Odelia Schwartz and Simoncelli, {Eero P.}",
year = "2001",
volume = "4",
pages = "819--825",
journal = "Nature Neuroscience",
issn = "1097-6256",
number = "8"}

@article{heeger1992normalization,
  title={Normalization of cell responses in cat striate cortex},
  author={Heeger, David J},
  journal={Visual Neuroscience},
  volume={9},
  number={2},
  pages={181--197},
  year={1992}}

@article{Peli90,
author = {Eli Peli},
journal = {J. Opt. Soc. Am. A},
number = {10},
pages = {2032--2040},
publisher = {Optica Publishing Group},
title = {Contrast in complex images},
volume = {7},
month = {Oct},
year = {1990}}

@inproceedings{Ortiz_2020_CVPR,
author = {Ortiz, Anthony and others},
title = {Local Context Normalization: Revisiting Local Normalization},
booktitle = {CVPR},
year = {2020}}

@inproceedings{ronneberger2015u,
title={U-net: Convolutional networks for biomedical image segmentation},
author={Ronneberger, Olaf and Fischer, Philipp and Brox, Thomas},
booktitle={MICCAI},
pages={234--241},
year={2015}}

@INPROCEEDINGS{raincityscapes,
author={Hu, Xiaowei and Fu, Chi-Wing and Zhu, Lei and Heng, Pheng-Ann},
booktitle={2019 CVPR}, 
title={Depth-Attentional Features for Single-Image Rain Removal}, 
year={2019},
pages={8014-8023}}

@misc { NormalizationPrinciplesinComputationalNeuroscience,
author = "Kenway Louie and Paul W. Glimcher",
title = "Normalization Principles in Computational Neuroscience",
year = "2019",
publisher = "Oxford University Press",
doi = "10.1093/acrefore/9780190264086.013.43"}





\end{document}